\documentclass{article}

\PassOptionsToPackage{numbers, compress}{natbib}

 \usepackage[preprint]{neurips_2026}

\usepackage[utf8]{inputenc} 
\usepackage[T1]{fontenc}    
\usepackage{hyperref}       
\usepackage{url}            
\usepackage{booktabs}       
\usepackage{amsfonts}       
\usepackage{nicefrac}       
\usepackage{microtype}      
\usepackage{xcolor}         
\usepackage{enumitem}
\usepackage{amsmath}
\usepackage{wrapfig}
\usepackage{xspace}
\newcommand{\ourmodel}{GeoFace\xspace}
\usepackage{graphicx}
\usepackage{multirow}
\usepackage{lipsum}
\usepackage{caption}
\usepackage{subcaption}

\title{GeoFace: Consistent Multi-View Face Generation \\ with Geometry-Constrained Diffusion}

\author{
\textbf{Yeji Choi}
\quad \textbf{Jinhyeok Choi}
\quad \textbf{Jaewon Min}
\quad \textbf{Minkyung Kwon}
\\[0.4em]
\textbf{Jin Hyeon Kim}
\quad \textbf{Seungryong Kim}$^{\dagger}$
\\[0.4em]
KAIST AI \qquad
\\[0.4em]
{\tt \href{https://github.com/cvlab-kaist/GeoFace/}{\textcolor{purple}{https://github.com/cvlab-kaist/GeoFace}}}
}

\begin{document}

\begingroup
\renewcommand{\thefootnote}{}
\footnotetext{$^\dagger$: Corresponding author}
\endgroup

\maketitle

\begin{figure}[h]
    \centering
   \vspace{-10pt}
   \includegraphics[width=\linewidth]{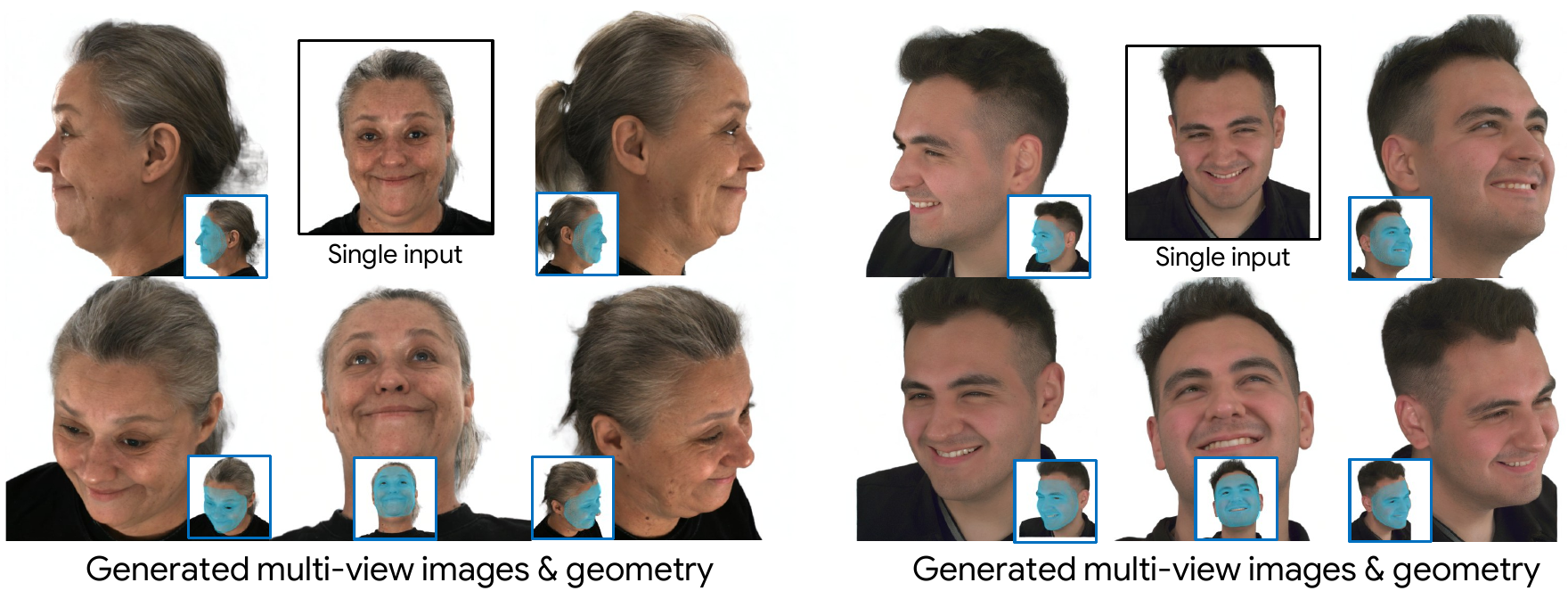}
    \vspace{-15pt}
    \caption{\textbf{GeoFace generates geometrically consistent multi-view images from a single input.} Given a single reference image, \ourmodel jointly generates multi-view facial images and 3D face geometry across diverse identities and viewpoints. The mesh overlay on each generated view demonstrates geometric alignment across viewpoints.}
    \label{fig:teaser}
\end{figure}

\begin{abstract}

We present \ourmodel, a geometry-constrained multi-view diffusion framework for consistent face generation from a single input.
While recent multi-view diffusion models achieve photorealistic synthesis at the per-view level, they lack an explicit mechanism to enforce a shared 3D structure across views, often leading to inconsistent geometry across viewpoints.
To address this, \ourmodel proposes a unified dual-stream framework for joint generation of multi-view RGB images and 3D face geometry, where the appearance and geometry streams interact through shared attention layers.
To encourage the two streams to mutually constrain each other, we introduce a geometry-guided attention alignment loss that supervises the cross-attention between appearance and geometry tokens with 3D-consistent correspondences, enabling the appearance stream to correctly reference pose-invariant geometric cues for robust alignment across viewpoints.
Geometry is represented as a canonical UV position map, derived from a FLAME mesh fitted to multi-view observations, serving as a view-invariant shared constraint across all generated views.
Experiments on RenderMe-360 and NeRSemble demonstrate that \ourmodel consistently outperforms existing methods in both visual quality and cross-view geometric consistency, facilitating more efficient 3D reconstruction.

\end{abstract}

\section{Introduction}
\label{sec:intro}

Generating photorealistic multi-view facial images has recently attracted significant interest in computer vision and graphics, driven by its broad applicability to 3D reconstruction, digital avatars, animation, and immersive content creation in virtual reality~\cite{Taubner2024CAP4DCA,gao2026high,lyu2025facelift, kocasari2026face_anything, xiang2024flashavatar}. In these applications, maintaining consistent geometry and identity under view changes is especially important for realistic rendering~\cite{kerbl20233dgs, mildenhall2021nerf} and coherent 3D understanding~\cite{qian2024gaussianavatars,xiang2024flashavatar}. 
Recent advances in deep generative models, particularly GANs~\cite{goodfellow2014generative, karras2019style} and diffusion models~\cite{song2020denoising, rombach2022high,blattmann2023align}, have led to remarkable progress in image and video synthesis, extending to highly photorealistic multi-view face generation~\cite{Gu2023DiffPortrait3DCD, Gu2025DiffPortrait360CP, Galanakis2025SpinMeRoundCM}. Despite these advances, generating geometrically consistent images from a single input image remains a challenging problem, as the limited field of view and lack of explicit 3D structural information make it difficult to recover a consistent underlying facial geometry across viewpoints.

The most common approaches rely on parametric face models such as 3DMMs~\cite{Blanz1999AMM_3dmm,booth20163d,booth2018large} and FLAME~\cite{Li2017LearningAM_flame} as geometric proxies for novel view synthesis. Combined with differentiable~\cite{khakhulin2022realistic,zheng2022avatar,zheng2023pointavatar,dib2021towards} or neural rendering techniques~\cite{wang2021one,hong2022headnerf,chen2024morphable,hu2017avatar}, these methods aim to maintain coarse geometric consistency.
Another line of work employs 3D-aware GANs~\cite{chan2022efficient_eg3d,An2023PanoHeadG3,Li2024SphereHeadS3,li2025hyplanehead,li2026condition_balancehead}, which implicitly encode geometry through tri-plane representations for controllable synthesis. However, fitting parametric models from a single image is inherently ill-posed, forcing the recovered geometry to converge toward a statistically averaged facial structure rather than faithfully preserving subject-specific geometry. As a result, both paradigms remain vulnerable to geometric artifacts and identity degradation, particularly under large pose variations.

More recently, diffusion-based multi-view generation methods~\cite{liu2023zero, shi2023zero123pp, liu2023syncdreamer, deng2023mv, voleti2024sv3d, Zhou2025StableVC, Gao2024CAT3DCA} have achieved compelling novel view synthesis through cross-view attention mechanisms and generalized camera conditioning. Building on these advances, face-specific methods~\cite{Gu2023DiffPortrait3DCD, Gu2025DiffPortrait360CP, Galanakis2025SpinMeRoundCM, Taubner2024CAP4DCA} further incorporate geometric cues such as head pose, expression, or normal maps~\cite{Taubner2024CAP4DCA, Galanakis2025SpinMeRoundCM} to improve facial consistency and structural fidelity. Although these efforts improve visual consistency, they still rely on view-specific geometric signals rather than a shared 3D representation, largely because geometrically aligned supervision across views has been overlooked in prior work. Consequently, recovering globally aligned facial geometry remains unresolved, despite its importance for faithfully preserving characteristic facial structures such as the nose, jawline, and forehead.

To address these limitations, we present \textbf{\ourmodel}, a diffusion framework that jointly generates multi-view RGB images and 3D facial geometry from a single reference image, enabling faithful preservation of subject-specific facial structures across views. Specifically, \ourmodel extends a multi-view diffusion backbone to jointly generate appearance and geometry within a unified framework, where the \textbf{appearance stream} and \textbf{geometry stream} interact through shared attention layers. Rather than treating geometry as an auxiliary signal, we integrate it directly into the generative process as an intrinsic representation shared across views. To further reinforce the interaction between the two streams, we introduce a geometry-guided attention alignment loss that explicitly supervises the cross-attention between geometry and appearance tokens using 3D correspondences derived from the canonical UV position map, which encodes dense 3D surface coordinates in a view-invariant parameterization. This encourages the geometry and appearance streams to mutually constrain each other, enabling the appearance stream to correctly reference pose-invariant geometric cues for robust synthesis of occluded or ambiguous regions, while grounding the geometry stream in the appearance context. 
Experiments demonstrate that \ourmodel achieves superior visual quality and significantly improved cross-view geometric consistency compared to existing multi-view diffusion methods that model appearance alone.
Experiments demonstrate that \ourmodel achieves superior visual quality and significantly improved cross-view geometric consistency compared to existing multi-view diffusion methods that model appearance alone. Furthermore, the jointly generated geometry accelerates downstream 3D reconstruction by providing a strong initialization prior. 

In summary, our main contributions are as follows:

\begin{itemize}[leftmargin=*]
\item We propose \ourmodel, a unified dual-stream diffusion framework where appearance and geometry streams interact through shared attention layers, enforcing cross-view geometric consistency as an intrinsic property of the generative process.
\item We introduce a geometry-guided attention alignment loss that supervises the cross-attention between geometry and appearance streams with 3D-consistent correspondences, encouraging the two streams to mutually constrain each other for geometrically consistent generation.
\item To facilitate shared geometry-constrained generation, we construct a canonical UV position map defined in FLAME space as a view-invariant geometry representation that is naturally compatible with diffusion models.
\item Extensive experiments demonstrate that \ourmodel achieves consistent improvements over existing baselines, with the jointly generated geometry further benefiting downstream 3D reconstruction.
\end{itemize}

\vspace{2mm}
\section{Related work}
\label{sec:rel_work}

\subparagraph{3D face reconstruction.}
3D face reconstruction aims to recover facial geometry from images, either as parametric shape and expression coefficients defined in 3DMMs~\cite{Blanz1999AMM_3dmm, booth20163d, booth2018large, Li2017LearningAM_flame} or as dense surface geometry~\cite{cao20133d, giebenhain2025pixel3dmm, taubner20243d_flowface}. Early optimization-based methods fit these models to 2D observations via photometric or landmark losses, while learning-based methods~\cite{deng2019accurate_deep3d, feng2021learning_deca, danvevcek2022emoca} improved efficiency by directly regressing model parameters from a single image. However, single-view reconstruction is inherently ill-posed, as a single image provides insufficient geometric cues to resolve depth ambiguity and recover occluded regions.
More recently, Pixel3DMM~\cite{giebenhain2025pixel3dmm} predicts dense pixel-aligned UV coordinates and surface normals via a DINOv2 backbone, enabling test-time FLAME optimization to recover fine-grained geometry. Transformer-based approaches such as VGGT~\cite{wang2025vggt} and Depth Anything 3~\cite{depthanything3} further improve geometric fidelity by leveraging multi-view cues through large-scale pretraining. Building on these advances, VGGTFace~\cite{ming2026vggtface} injects topology into VGGT's point map representation using pixel-aligned UV values from Pixel3DMM and refines the result via topology-aware bundle adjustment to produce topologically consistent facial meshes. 

\subparagraph{Multi-view diffusion models.}
Diffusion-based novel view synthesis has advanced rapidly, with Zero123~\cite{liu2023zero} pioneering single-image-to-novel-view generation by conditioning on relative camera poses. Subsequent works such as SyncDreamer~\cite{liu2023syncdreamer}, Zero123++~\cite{shi2023zero123pp}, and MVDiffusion~\cite{deng2023mv} extend this to simultaneous multi-view generation with improved cross-view consistency, while SV3D~\cite{voleti2024sv3d} and SEVA~\cite{Zhou2025StableVC} leverage video diffusion priors to enforce temporal coherence across views. CAT3D~\cite{Gao2024CAT3DCA} generalizes this paradigm by enabling reference-based multi-view generation from arbitrary camera configurations, which we adopt as our backbone. However, these methods focus primarily on appearance synthesis and lack explicit geometric awareness, often suffering from multi-view geometric inconsistencies analogous to the Janus problem observed in text-to-3D generation. While cross-view attention mechanisms implicitly encourage feature-level consistency, they provide no guarantee of a coherent underlying 3D structure, limiting their applicability to downstream tasks such as 3D reconstruction that require accurate cross-view correspondence.

\subparagraph{Multi-view face generation.}
Early 3D-aware face generation methods leverage generative adversarial networks (GANs)~\cite{goodfellow2014generative, karras2019style, karras2020analyzing} combined with implicit neural representations to synthesize photorealistic faces with explicit camera control~\cite{szabo2019unsupervised, shi2021lifting, chan2022efficient_eg3d, An2023PanoHeadG3, Li2024SphereHeadS3, li2025hyplanehead}. EG3D~\cite{chan2022efficient_eg3d} introduces a tri-plane representation that efficiently factorizes 3D features for neural rendering, and subsequent works~\cite{An2023PanoHeadG3, Li2024SphereHeadS3, li2025hyplanehead, li2026condition_balancehead} extend this to 360-degree synthesis and address artifacts under extreme viewpoints through alternative coordinate representations. Although these methods achieve impressive visual quality, they lack explicit geometric constraints, leading to identity degradation and geometric artifacts under large viewpoint changes.
More recently, diffusion-based methods adapt multi-view generation to the portrait domain. DiffPortrait3D~\cite{Gu2023DiffPortrait3DCD} and DiffPortrait360~\cite{Gu2025DiffPortrait360CP} introduce identity-preserving portrait generation with 3D-aware noise conditioning, extending to 360-degree synthesis. SpinMeRound~\cite{Galanakis2025SpinMeRoundCM} improves rotational consistency using camera-aware diffusion priors, and CAP4D~\cite{Taubner2024CAP4DCA} incorporates FLAME-based parametric face models as geometric guidance within the diffusion process. Despite compelling visual quality, these methods remain appearance-driven with limited mechanisms to enforce a shared 3D geometry across views.

\begin{figure}[t]
  \centering
  \includegraphics[width=\textwidth]{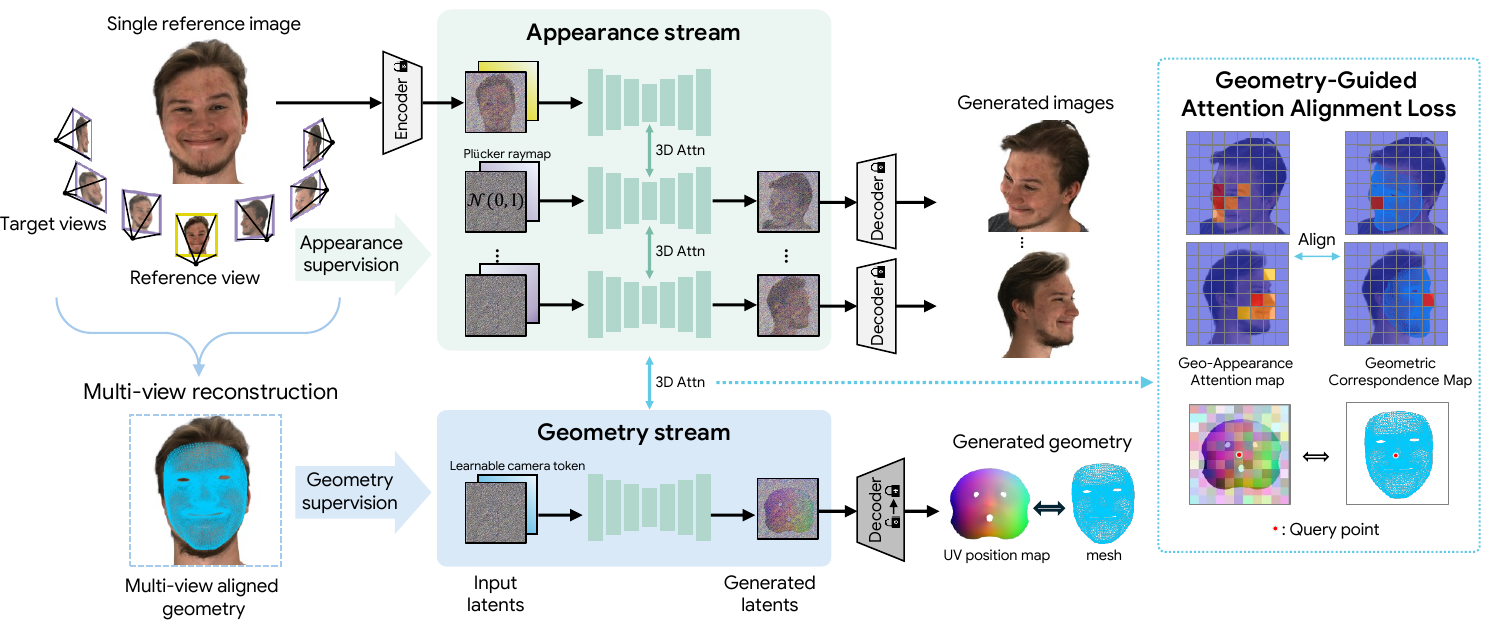}
\caption{\textbf{Overview of GeoFace.} Given a single reference image and target camera poses, \ourmodel jointly generates multi-view RGB images and 3D face geometry within a unified dual-stream framework. The \textbf{appearance stream} denoises target view latents conditioned on Plücker ray embeddings via shared 3D attention layers, while the \textbf{geometry stream} denoises a geometry latent conditioned on a learnable camera token. Both streams are supervised by ground-truth multi-view images and multi-view aligned geometry, respectively. Consequently, the jointly generated images share a consistent underlying 3D structure across all viewpoints.}
\label{fig:overview}
\end{figure}

\section{Method}
\label{sec:method}

\subsection{Overview}
An overview of \ourmodel is shown in Fig.~\ref{fig:overview}. Given a reference image and target camera poses, \ourmodel produces a set of target views alongside a shared canonical UV position map that encodes the underlying 3D face geometry as a view-invariant constraint across all generated views.
\ourmodel extends a multi-view diffusion backbone with a dual-stream architecture. The \textbf{appearance stream} denoises target view latents conditioned on Plücker ray embeddings, while the \textbf{geometry stream} jointly denoises a geometry latent representing the canonical UV position map. Both streams interact through shared 3D attention layers, allowing appearance and geometry to mutually constrain each other during the denoising process, enforcing a coherent 3D structure across all generated views. In the following subsections, we detail the model architecture, geometry stream design, and dataset construction for geometry supervision.

\paragraph{Problem formulation.}
We formulate multi-view face generation as a conditional joint generation problem over appearance and geometry from a single image using a latent diffusion model.
Given a reference image $I^{\text{ref}} \in \mathbb{R}^{H \times W \times 3}$ and its estimated camera parameters $c^{\text{ref}}$, we define a set of $N-1$ target camera conditions $\mathbf{C}^{\text{tgt}} = \{ c_{(v)}^{\text{tgt}} \}_{v=1}^{N-1}$ based on relative transformations with respect to $c^{\text{ref}}$.
Our goal is to jointly generate $N-1$ multi-view target RGB images $\mathbf{I}^{\text{tgt}} = \{ I_{(v)}^{\text{tgt}} \}_{v=1}^{N-1}$ and a shared geometry representation $\mathcal{G}$ in the FLAME canonical space, by modeling the joint distribution:
\begin{equation}
    P(\mathbf{I}^{\text{tgt}}, \mathcal{G} \mid I^{\text{ref}}, c^{\text{ref}}, \mathbf{C}^{\text{tgt}}),
\end{equation}
where $\mathcal{G}$ denotes a canonical UV position map that encodes dense 3D surface coordinates shared across all generated views.
By jointly modeling $\mathcal{G}$ as a global canonical representation coupled with appearance generation, each RGB view is implicitly constrained by a common underlying 3D structure, allowing geometry to act as an explicit consistency prior during the diffusion process.

\subsection{Dual-Stream Multi-View Diffusion}
\label{subsec:method_diffusion} 
\ourmodel extends a multi-view diffusion backbone with a dual-stream architecture that jointly denoises appearance and geometry latents through shared attention layers, enabling geometry to act as an explicit cross-view consistency constraint during the generative process.

\paragraph{Model architecture.}
\ourmodel builds upon a CAT3D-style latent multi-view diffusion backbone~\cite{Gao2024CAT3DCA}, which jointly denoises $N$ streams within a single U-Net by processing all views simultaneously through shared attention layers. 
We take as input the reference image $I^{\text{ref}}$ along with $N$ noisy target latents, and condition each view on its corresponding camera pose encoded as a 6-channel Pl\"ucker embedding aligned with the latent resolution.
We extend this backbone to support dual-stream generation by repurposing the last generation stream to produce geometry in place of an RGB view. 
While the first $N-1$ streams generate target RGB views $\{I_{(v)}^{\text{tgt}}\}_{v=1}^{N-1}$, the final stream produces a shared geometry representation $\mathcal{G}$, encoded as a UV position map—a structured $H \times W \times 3$ grid where each texel stores the 3D canonical coordinate of the corresponding surface point on the FLAME mesh.
This image-like representation preserves spatial structure, making it directly compatible with 2D convolutional architectures without requiring architectural modification.

\paragraph{Conditions.}
To provide stable identity and appearance conditioning, we select a near-frontal image as the reference view $I^{\text{ref}}$. 
A set of target camera conditions $\mathbf{C}^{\text{tgt}}$ is defined relative to the reference camera pose by specifying yaw and pitch offsets. 
This camera formulation makes the model agnostic to the absolute world coordinate system, enabling generalization across diverse camera configurations at inference time.
We further assign a conditioning mask $\mathbf{m}$ to encode the role of each stream, where values of $0$, $1$, and $0.5$ correspond to the reference, target, and geometry streams, respectively.

\begin{figure*}[t]
\centering

\begin{subfigure}[t]{0.55\linewidth}
\vspace{0pt}
    \centering
    \includegraphics[width=\linewidth]{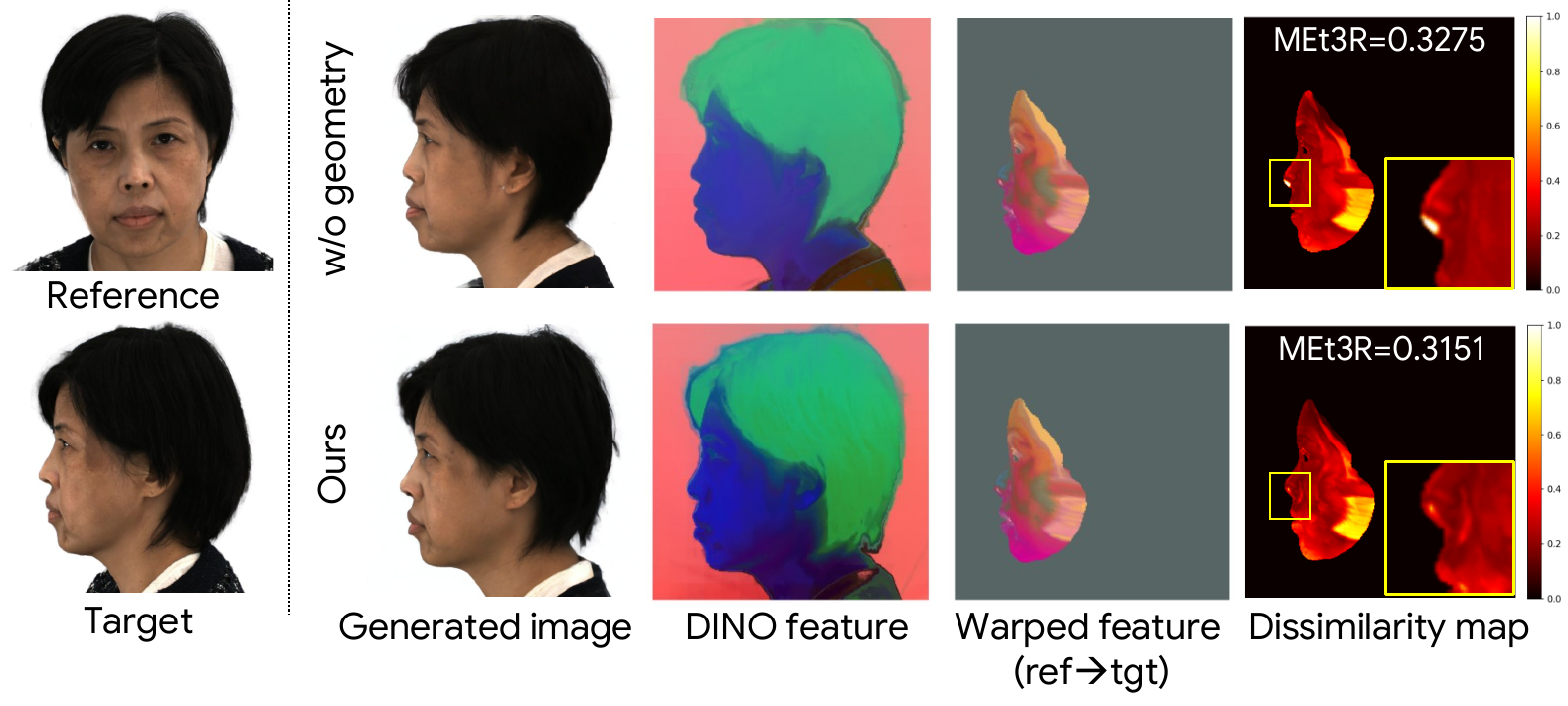}
    \caption{Cross-view consistency visualization.}
\end{subfigure}
\hspace{0.01\linewidth}
\begin{subfigure}[t]{0.4\linewidth}
\vspace{30pt}
    \centering
    \resizebox{\linewidth}{!}{
        \begin{tabular}{lccc}
        \toprule
        Method & w/o geometry & Ours & GT \\
        \midrule
        MEt3R($\downarrow$) & 0.3304 & 0.3198 & 0.3157 \\
        \bottomrule
        \end{tabular}
    }
    \caption{Quantitative MEt3R comparison.}
\end{subfigure}

\caption{
\textbf{Cross-view feature consistency analysis.} We compare GeoFace against its variant without the geometry stream using MEt3R~\cite{asim2025met3r}, both qualitatively~(a) and quantitatively~(b). Results in~(b) are averaged over 40 test identities on RenderMe-360~\cite{Pan2023RenderMe360AL}. Lower MEt3R indicates better consistency.}
\label{fig:analysis}

\end{figure*}

\subsection{Geometry Stream Design}
\label{sec:method_geometry}

\paragraph{Motivation.} Multi-view diffusion models trained with appearance-only objectives implicitly learn cross-view correspondences through shared attention, yet lack an explicit mechanism to enforce a geometrically coherent 3D structure. To quantify this, we analyze cross-view feature consistency using MEt3R~\cite{asim2025met3r}, computing scores between the input reference image and each generated novel-view image; lower MEt3R indicates better geometric consistency. As shown in Fig.~\ref{fig:analysis}, the appearance-only model (w/o geometry stream) yields higher dissimilarity, particularly at facial boundary regions, revealing that appearance-only generation fails to recover a consistent underlying 3D surface. This motivates our dedicated geometry stream, which jointly denoises a shared geometry representation alongside the appearance streams, encouraging the model to learn a common underlying 3D structure across all views by exposing explicit geometric tokens to the shared attention layers.

\paragraph{Geometry representation.}
We represent 3D face geometry as a canonical UV position map $\mathcal{G} \in \mathbb{R}^{H \times W \times 3}$, where each texel encodes the dense 3D surface coordinate of the corresponding point on the FLAME mesh.
This representation is topology-aware and view-invariant by construction, as it is defined in FLAME canonical space rather than any particular camera viewpoint.
Critically, its image-like spatial structure makes it directly compatible with the 2D convolutional architecture of the diffusion backbone, enabling joint denoising with RGB streams through shared attention layers without any architectural modification.
Furthermore, since the pretrained SD VAE encoder is sufficiently expressive for representing UV maps, we keep it frozen and fine-tune only the decoder to handle the distribution mismatch arising from sharp discontinuities along UV seam boundaries.

\paragraph{Learnable camera token.}
Since $\mathcal{G}$ is defined in FLAME canonical space and does not correspond to any particular camera viewpoint, applying a standard Pl\"ucker ray embedding as the camera condition would impose a spurious viewpoint assumption.
We therefore replace it with a learnable token $\mathbf{e}^{\text{geo}} \in \mathbb{R}^{6}$, optimized end-to-end as an implicit positional embedding for canonical geometry.

\paragraph{Geometry-Guided Attention Alignment Loss.}
The geometry stream introduces a cross-modal attention pathway between UV geometry tokens and appearance tokens that receives no correspondence signal during pretraining, leaving the geometric consistency of this pathway unsupervised.
To explicitly enforce 3D-consistent cross-modal attention, we propose a geometry-guided attention alignment loss that supervises this pathway using geometric correspondences derived from the FLAME UV position map and calibrated camera poses, applied bidirectionally to both geometry-to-appearance and appearance-to-geometry cross-attention.

The geometry-guided attention alignment loss is defined as the bidirectional cross-entropy between the predicted geometry-to-appearance attention $\hat{A}^{\text{uv}} \in \mathbb{R}^{hw \times F \cdot hw}$ and appearance-to-geometry attention $\hat{A}^{\text{face}} \in \mathbb{R}^{F \cdot hw \times hw}$ and their respective GT correspondences:
\begin{equation}
    \mathcal{L}_{\text{align}} = 
    -\sum_{i \in \mathcal{V}^{\text{uv}}} \sum_{v,j} \mathbf{g}^{\text{uv}}_{i,(v,j)} \log \hat{A}^{\text{uv}}_{i,(v,j)}
    -\sum_{(v,j) \in \mathcal{V}^{\text{face}}} \sum_{i} \mathbf{g}^{\text{face}}_{(v,j),i} \log \hat{A}^{\text{face}}_{(v,j),i},
\end{equation}
where $h \times w$ denotes the spatial token resolution of the latent features, $\mathcal{V}^{\text{uv}}$ and $\mathcal{V}^{\text{face}}$ denote valid query tokens with at least one visible correspondence, $v \in \{1, \ldots, F\}$ denotes the face view index, and $j$ denotes the spatial token index within each view. For each UV latent token $i$ and face-view latent token $(v, j)$, their 3D positions $\mathbf{p}^{\text{uv}}_i$ and $\mathbf{p}^{\text{face}}_{v,j}$ are obtained by averaging the world-space coordinates of foreground UV texels within the corresponding latent cell.
The ground-truth correspondence $\mathbf{g}^{\text{uv}}$ is defined as an argmax one-hot over pairwise L2 distances between $\mathbf{p}^{\text{uv}}$ and $\mathbf{p}^{\text{face}}$, with pairs exceeding $\tau$ masked to suppress spurious matches, and $\mathbf{g}^{\text{face}}$ is approximated as its transpose.

This bidirectional supervision reinforces the mutual constraint between geometry and appearance streams, encouraging geometrically consistent cross-modal attention.
As face-to-face cross-attention already exhibits well-aligned correspondences in the portrait domain and incurs quadratic memory overhead, we restrict correspondence supervision to the geometry-appearance pathway where explicit geometric guidance is most needed.

\paragraph{Training.}
The model is trained jointly over all $N$ streams with the following objective:
\begin{equation}
    \mathcal{L}_{\text{total}} = \mathcal{L}_{\text{rgb}} + \mathcal{L}_{\text{geo}} + \lambda_{\text{align}}\mathcal{L}_{\text{align}},
\end{equation}
where $\mathcal{L}_{\text{rgb}}$ is the v-prediction loss over the $N-1$ appearance streams, $\mathcal{L}_{\text{geo}}$ is the v-prediction loss over the geometry stream, and $\mathcal{L}_{\text{align}}$ is the geometry-guided attention alignment loss weighted by $\lambda_{\text{align}}$.

\subsection{Dataset Construction for Geometry Supervision}
\label{subsec:method_dataset}
Since canonicalized multi-view geometry supervision is not directly available in existing datasets, we construct it from multi-view face video collections, including RenderMe-360~\cite{Pan2023RenderMe360AL} and Nersemble v2~\cite{kirschstein2023nersemble}. From each multi-view sequence, we sample frames with noticeable facial expressions to capture diverse geometry variations while preserving strict synchronization across views. RenderMe-360 provides 27 synchronized viewpoints per identity and Nersemble v2 provides 16, yielding dense multi-view observations suitable for accurate geometry reconstruction.

\paragraph{Multi-view reconstruction and canonicalization.}
We recover 3D geometry from sampled views using the VGGTFace pipeline~\cite{ming2026vggtface}, which aggregates per-view point maps from VGGT~\cite{wang2025vggt} and UV coordinate maps from Pixel3DMM~\cite{giebenhain2025pixel3dmm} across all views, and jointly refines them via bundle adjustment~\cite{triggs1999bundle} to produce a consistent set of FLAME vertices. Since the resulting vertices are defined in an arbitrary world coordinate system, we apply a Sim(3) transformation estimated via Procrustes alignment~\cite{Umeyama1991LeastSquaresEO} to map all meshes into a shared FLAME canonical space, removing scale and pose ambiguity and ensuring consistent geometry across identities.

\paragraph{Geometry supervision.}
The canonicalized meshes are then rasterized into the UV domain defined by the FLAME topology, yielding a single UV position map per frame that encodes dense 3D surface coordinates at each texel. By fitting to multi-view observations and canonicalizing into FLAME space, this representation serves as a view-invariant geometry supervision signal for training the geometry stream.

\section{Experiments}
\label{sec:exp}

\subsection{Experimental settings}

\paragraph{Implementation details.}
We adopt CAT3D~\cite{Gao2024CAT3DCA} as the base multi-view diffusion model and initialize it from pretrained Stable Diffusion 2.1 weights~\cite{rombach2022high}, following prior work~\cite{Gao2024CAT3DCA, Galanakis2025SpinMeRoundCM}. As the official implementation of CAT3D is not publicly available, we implement our method using the MVGenMaster~\cite{Cao2024MVGenMasterSM} codebase.
For training, we use $N=7$ target views per sample, resulting in 8 streams in total (1 input view, 6 target views, and 1 UV position map). The model is optimized using AdamW~\cite{Loshchilov2017DecoupledWD} with a learning rate of $5 \times 10^{-5}$.
The per-GPU batch size is 6, and the model is trained for 50k iterations on 2 NVIDIA A100 (40GB) GPUs. We apply gradient clipping (max norm 1.0) and randomly drop conditioning inputs with a probability of 0.1 to enable classifier-free guidance (CFG)~\cite{Ho2022ClassifierFreeDG}. At inference, we use a guidance scale of 2.0 and adopt the DDIM sampler~\cite{song2020denoising} with 50 sampling steps. All experiments are conducted at a resolution of $512 \times 512$.
For the geometry-guided attention alignment loss, we follow CAMEO~\cite{kwon2025cameo} and apply supervision at layer 10 of the UNet decoder with $\lambda_{\text{align}} = 0.02$ and correspondence distance threshold $\tau = 0.035$.

\paragraph{Datasets.}
We train our model using two multi-view video datasets: RenderMe-360~\cite{Pan2023RenderMe360AL} and Nersemble v2~\cite{kirschstein2023nersemble}.
RenderMe-360 contains 500 identities, each recorded with 12 expressions across 60 synchronized viewpoints covering the full head. As we focus on face-centric multi-view generation, we restrict the viewpoints to those within $\pm90^\circ$ of the frontal direction. For training, we randomly select one representative frame per sequence that reflects the corresponding expression. We use 450 identities for training and 40 for testing, yielding over 5K training pairs and 480 testing pairs.
Nersemble v2 contains multi-view facial sequences captured from 16 viewpoints with consistent camera calibration. Following the same protocol, we randomly select one representative frame per sequence and use 450 identities for training and 30 for testing. For each identity, we sample 15 view pairs, resulting in 9,750 training pairs and 450 testing pairs.

\begin{table*}[t]
\caption{\textbf{Quantitative comparisons of novel-view synthesis on RenderMe-360~\cite{Pan2023RenderMe360AL}.} We evaluate performance across two viewpoint ranges: frontal-to-mid views (up to $\pm45^\circ$) and profile views ($\pm45^\circ$ to $\pm90^\circ$).}

\vspace{2mm}
\label{tab:quan_renderme}
\centering
\resizebox{\linewidth}{!}{
\begin{tabular}{clcccccccc}

\toprule
\multirow{2}{*}{Category} & \multirow{2}{*}{Method}
& \multicolumn{4}{c}{Frontal-to-mid views}
& \multicolumn{4}{c}{Profile views} \\

& 
& PSNR$\uparrow$ & SSIM$\uparrow$ & LPIPS$\downarrow$ & CSIM$\uparrow$
& PSNR$\uparrow$ & SSIM$\uparrow$ & LPIPS$\downarrow$ & CSIM$\uparrow$ \\
\midrule

\multirow{2}{*}{\begin{tabular}[c]{@{}c@{}}NeRF\\-based\end{tabular}}
& PanoHead~\cite{An2023PanoHeadG3}
& 11.78 & 0.6404 & 0.3957 & 0.5330
& 10.71 & 0.5985 & 0.4622 & 0.3754 \\

& SphereHead~\cite{Li2024SphereHeadS3}
& 12.24 & 0.6528 & 0.3883 & 0.5219
& 11.29 & 0.6252 & 0.4471 & 0.3601 \\

\midrule

\multirow{4}{*}{\begin{tabular}[c]{@{}c@{}}Diffusion\\-based\end{tabular}}
& SEVA~\cite{Zhou2025StableVC}
& 12.39 & 0.6706 & 0.3568 & 0.5895
& 10.26 & 0.6228 & 0.4650 & 0.4848 \\

& DiffPortrait360~\cite{Gu2025DiffPortrait360CP}
& 10.59 & 0.5915 & 0.4534 & 0.4697
& 8.11 & 0.4804 & 0.6005 & 0.2513 \\

& CAP4D~\cite{Taubner2024CAP4DCA}
& 12.82 & 0.6782 & 0.3235 & 0.6526
& 12.10 & 0.6676 & 0.3738 & 0.5388 \\

& \textbf{GeoFace (Ours)}
& \textbf{17.34} & \textbf{0.7780} & \textbf{0.1795} & \textbf{0.8084}
& \textbf{15.30} & \textbf{0.7562} & \textbf{0.2440} & \textbf{0.7001} \\

\bottomrule
\end{tabular}
}
\end{table*}

\begin{table*}[t]                        
\caption{\textbf{Quantitative comparisons of novel-view synthesis on Nersemble v2~\cite{kirschstein2023nersemble}.} We evaluate performance
across two viewpoint ranges: frontal-to-mid views (up to $\pm45^\circ$) and profile views ($\pm45^\circ$ to $\pm90^\circ$).}                
\vspace{2mm}                              
\label{tab:quan_nersemble}          
\centering                               
\resizebox{\linewidth}{!}{
\begin{tabular}{clcccccccc}            
\toprule
\multirow{2}{*}{Category} & \multirow{2}{*}{Method}    
& \multicolumn{4}{c}{Frontal-to-mid views}   
& \multicolumn{4}{c}{Profile views} \\   
&                                     
& PSNR$\uparrow$ & SSIM$\uparrow$ & LPIPS$\downarrow$ & CSIM$\uparrow$           
& PSNR$\uparrow$ & SSIM$\uparrow$ & LPIPS$\downarrow$ & CSIM$\uparrow$ \\   
\midrule                              
\multirow{2}{*}{\begin{tabular}[c]{@{}c@{}}NeRF\\-based\end{tabular}}  
& PanoHead~\cite{An2023PanoHeadG3}
& 13.54 & 0.6613 & 0.3528 & 0.5983                
& 11.80 & 0.6310 & 0.4358 & 0.5088 \\            
& SphereHead~\cite{Li2024SphereHeadS3}    
& 13.44 & 0.6650 & 0.3556 & 0.6157       
& 11.35 & 0.6263 & 0.4433 & 0.5206 \\     
\midrule                       
\multirow{4}{*}{\begin{tabular}[c]{@{}c@{}}Diffusion\\-based\end{tabular}} 

& SEVA~\cite{Zhou2025StableVC}
& 12.73 & 0.2169 & 0.4692 & 0.5436
& 11.18 & 0.2073 & 0.4849 & 0.4839 \\

& DiffPortrait360~\cite{Gu2025DiffPortrait360CP}
& 12.31 & 0.6250 & 0.4064 & 0.5823
& 10.67 & 0.5803 & 0.4927 & 0.4795 \\

& CAP4D~\cite{Taubner2024CAP4DCA}
& 12.41 & 0.6428 & 0.4059 & 0.6090
& 10.68 & 0.6006 & 0.5099 & 0.4606 \\


& \textbf{GeoFace (Ours)}
& \textbf{21.33} & \textbf{0.7858} & \textbf{0.1460} & \textbf{0.8365}
& \textbf{18.16} & \textbf{0.7384} & \textbf{0.2055} & \textbf{0.7782} \\

\bottomrule
\end{tabular}
}
\end{table*}

\paragraph{Baselines.}
We compare our method against two groups of baselines. Face-specific 3D-aware generative models include PanoHead~\cite{An2023PanoHeadG3} and SphereHead~\cite{Li2024SphereHeadS3}, which are NeRF-based approaches, as well as DiffPortrait360~\cite{Gu2025DiffPortrait360CP} and CAP4D~\cite{Taubner2024CAP4DCA}, which are diffusion-based methods. These methods are designed for high-quality head reconstruction and synthesis. For CAP4D, we report results using only the morphable multi-view diffusion model (MMDM) without the 3DGS stage for fair comparison of novel-view synthesis quality. General multi-view or video generation models include SEVA~\cite{Zhou2025StableVC}, which aims to produce consistent outputs across views without explicit face priors. All methods are evaluated on the same test splits of RenderMe-360 and Nersemble v2 for fair comparison.

\paragraph{Evaluation metrics.}
We evaluate performance in terms of image quality, identity consistency, and geometric alignment. For image quality, we report PSNR and SSIM to measure pixel-level and structural fidelity (higher is better), and LPIPS~\cite{zhang2018unreasonable} to assess perceptual similarity (lower is better). To assess identity consistency across views, we compute cosine similarity (CSIM) between identity features extracted using a pretrained face recognition model ArcFace~\cite{deng2019arcface} (higher is better).

\begin{figure}[t]
  \centering
  \includegraphics[width=\textwidth]{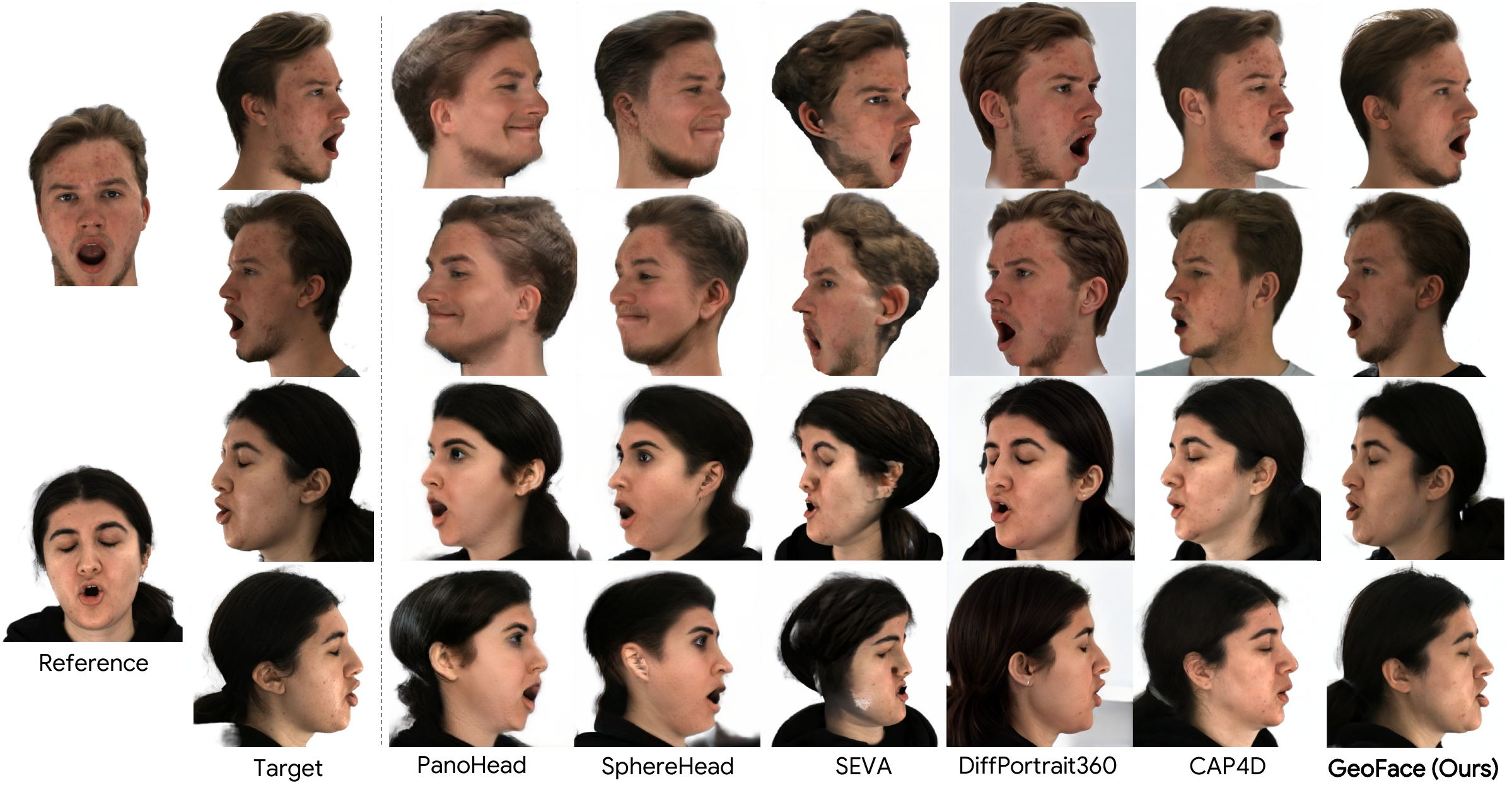}
\caption{\textbf{Qualitative comparisons of novel-view synthesis on RenderMe-360~\cite{Pan2023RenderMe360AL}.} For each identity, we show the reference image alongside generated profile views from all baselines and our method.}
\label{fig:main_qual}
\end{figure}

\begin{figure}[t]
  \centering
  \includegraphics[width=\textwidth]{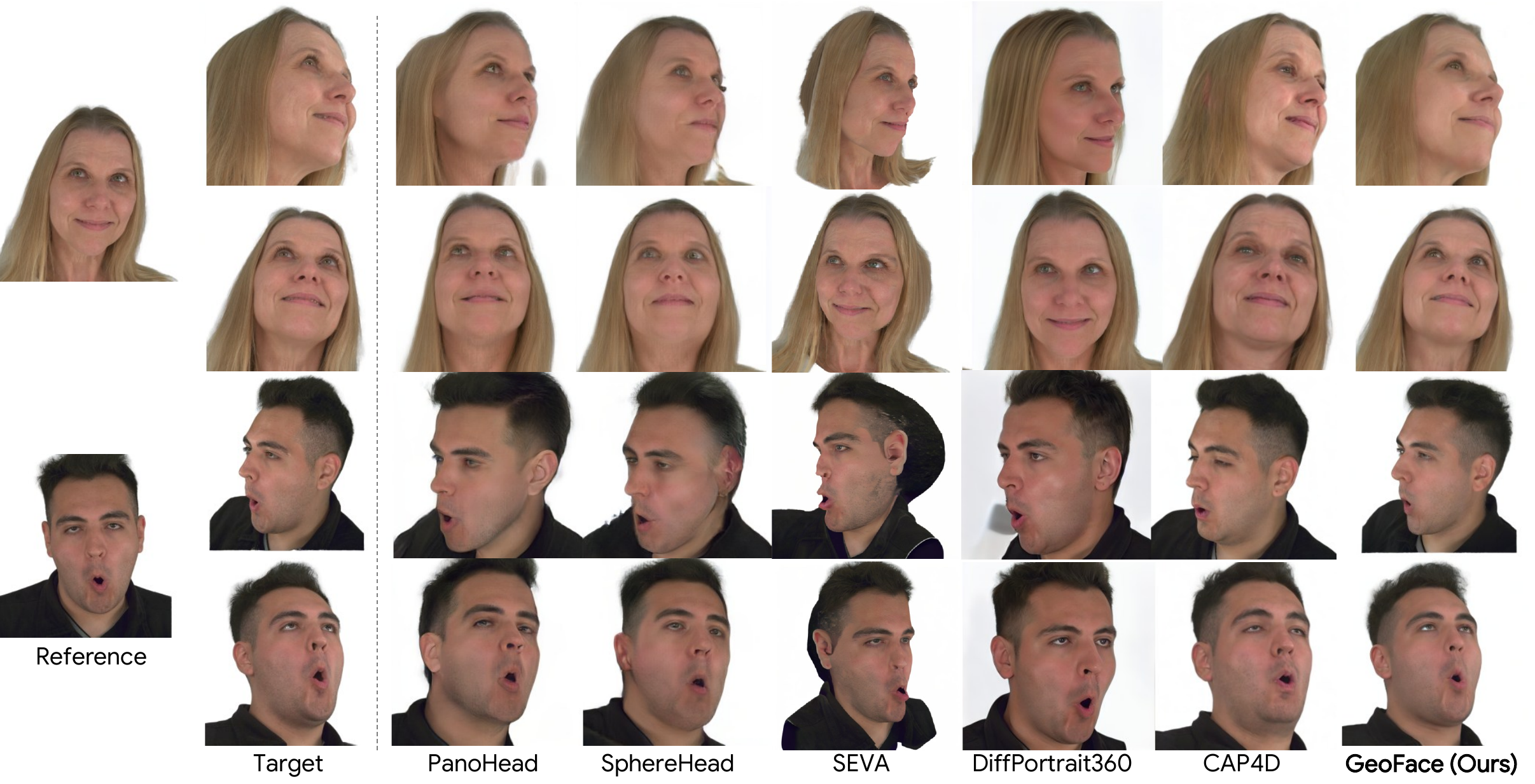}
\caption{\textbf{Qualitative comparisons of novel-view synthesis on Nersemble v2~\cite{kirschstein2023nersemble}.} For each identity, we show the reference image alongside generated profile views from all baselines and our method.}
\label{fig:main_qual2}
\end{figure}

\subsection{Experimental results}
\label{sec:exp_result}

\paragraph{Quantitative results.}
Tables~\ref{tab:quan_renderme} and \ref{tab:quan_nersemble} report quantitative comparisons on RenderMe-360 and Nersemble, respectively. \ourmodel consistently outperforms all baselines across both datasets, both viewpoint ranges, and all evaluation metrics. The improvement in CSIM confirms that jointly modeling geometry and appearance leads to better identity preservation across viewpoints. Notably, the gains are more pronounced in profile views, demonstrating that explicit geometry supervision is especially beneficial under large pose variations, where appearance-only methods tend to degrade. NeRF-based methods and face-specific diffusion methods trail significantly behind \ourmodel, highlighting the advantage of incorporating explicit geometric supervision within a strong generative diffusion prior.

\paragraph{Qualitative results.}
Fig.~\ref{fig:main_qual} and Fig.~\ref{fig:main_qual2} present qualitative comparisons on RenderMe-360 and Nersemble v2, respectively. NeRF-based methods, PanoHead and SphereHead, struggle to synthesize realistic facial appearance under large pose variations, possibly due to the limited diversity of large-pose training data. SEVA, trained on general objects and scenes, lacks understanding of facial structure, resulting in severely distorted head shapes and facial features such as the nose and forehead in novel views. DiffPortrait360 produces visually appealing results but exhibits geometric inconsistencies under large pose variations, particularly around the nose and jaw regions. CAP4D conditions generation on FLAME fitting parameters, but when the fitting fails to capture fine-grained expressions or geometry, the generated views suffer from structural inconsistencies. 
In contrast, \ourmodel produces geometrically consistent and identity-preserving results even under large pose variations. The improvement is particularly pronounced at self-occluded regions such as the jaw and nasal bridge, where appearance-only models lack sufficient constraints to resolve depth ambiguity—regions where the geometry stream provides the most direct supervisory signal.

\begin{figure}[t]
  \centering
  \includegraphics[width=\textwidth]{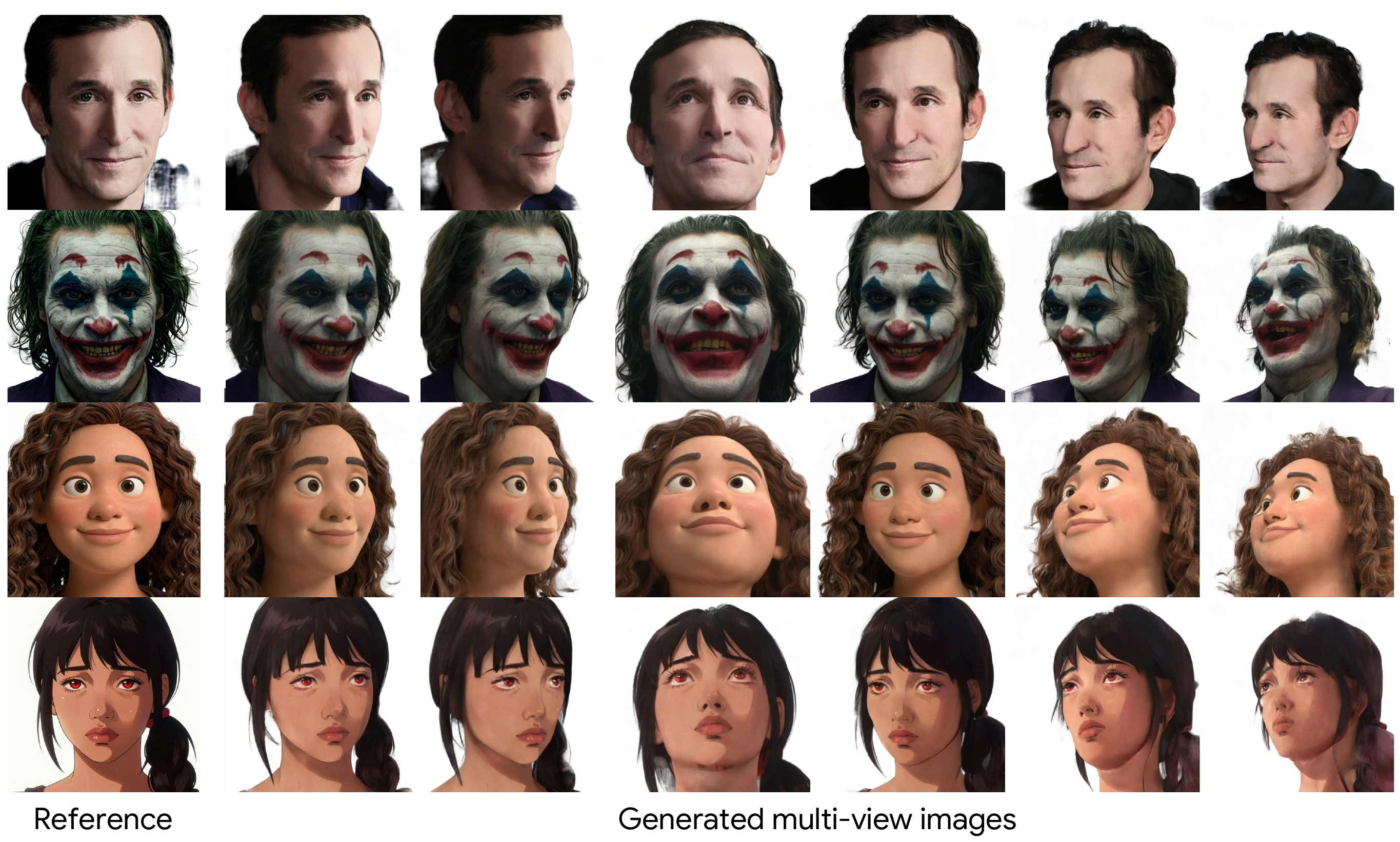}
\caption{\textbf{In-the-wild results.} GeoFace generalizes to diverse input types including portraits under challenging lighting, heavily made-up faces, 3D-rendered characters, and stylized illustrations. }
\label{fig:qual_wild}
\end{figure}

\begin{figure}[h]
  \centering
  \includegraphics[width=\linewidth]{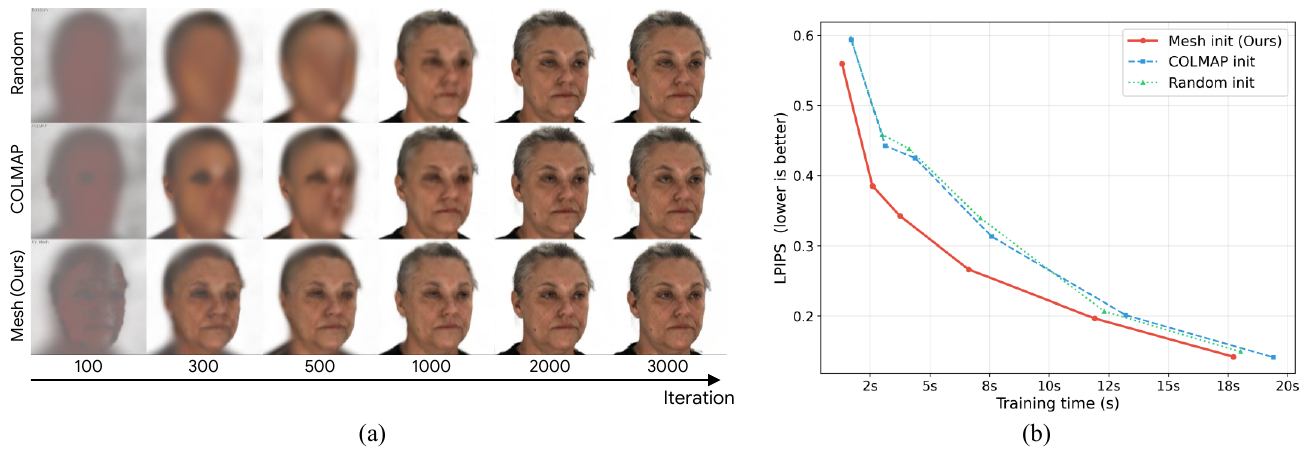}
\caption{\textbf{Downstream 3D Gaussian Splatting reconstruction.} (a) Qualitative comparison of reconstruction quality across training iterations under three initialization strategies. (b) LPIPS convergence curves over training time. Mesh-based initialization using the jointly generated FLAME mesh achieves faster convergence and lower LPIPS throughout training compared to random and COLMAP-based initialization.}
\label{fig:GS_recon}
\end{figure}

\paragraph{In-the-wild results.}
Fig.~\ref{fig:qual_wild} demonstrates the generalization capability of \ourmodel beyond controlled capture settings. Across diverse input types — including portraits under challenging lighting conditions, heavily made-up faces, 3D-rendered characters, and stylized illustrations — \ourmodel consistently produces geometrically coherent novel views while preserving the distinctive appearance of each input. These results suggest that the geometry stream learns a domain-agnostic structural prior, enabling robust multi-view synthesis even under significant domain shifts from the training distribution.

\paragraph{Downstream 3D reconstruction.}
To validate the usefulness of the jointly generated geometry, we evaluate its effectiveness as an initialization prior for 3D Gaussian Splatting~\cite{kerbl20233dgs}. Specifically, we generate 24 multi-view images from a single reference image using GeoFace and reconstruct a 3D Gaussian representation from these views. For initialization, we use the averaged FLAME mesh across multiple generations as the initialization geometry. We compare three initialization strategies: random initialization, COLMAP-based initialization, and our mesh-based initialization. Since COLMAP relies on sparse feature matching, the reconstructed points are typically concentrated around highly textured regions such as the eyes, resulting in limited coverage of the facial surface. In contrast, our method directly leverages the generated mesh to provide dense and uniform Gaussian point placement over the entire face.
As shown in Figure~\ref{fig:GS_recon}(a), our mesh-based initialization produces noticeably sharper reconstructions during the early stages of optimization. Figure~\ref{fig:GS_recon}(b) further confirms faster convergence and consistently lower LPIPS compared to random and COLMAP-based initialization, demonstrating that the jointly generated geometry serves as an effective structural prior for downstream 3D reconstruction.

\subsection{Ablation Study}
\begin{figure}[t]
  \centering
  \includegraphics[width=0.85\textwidth]{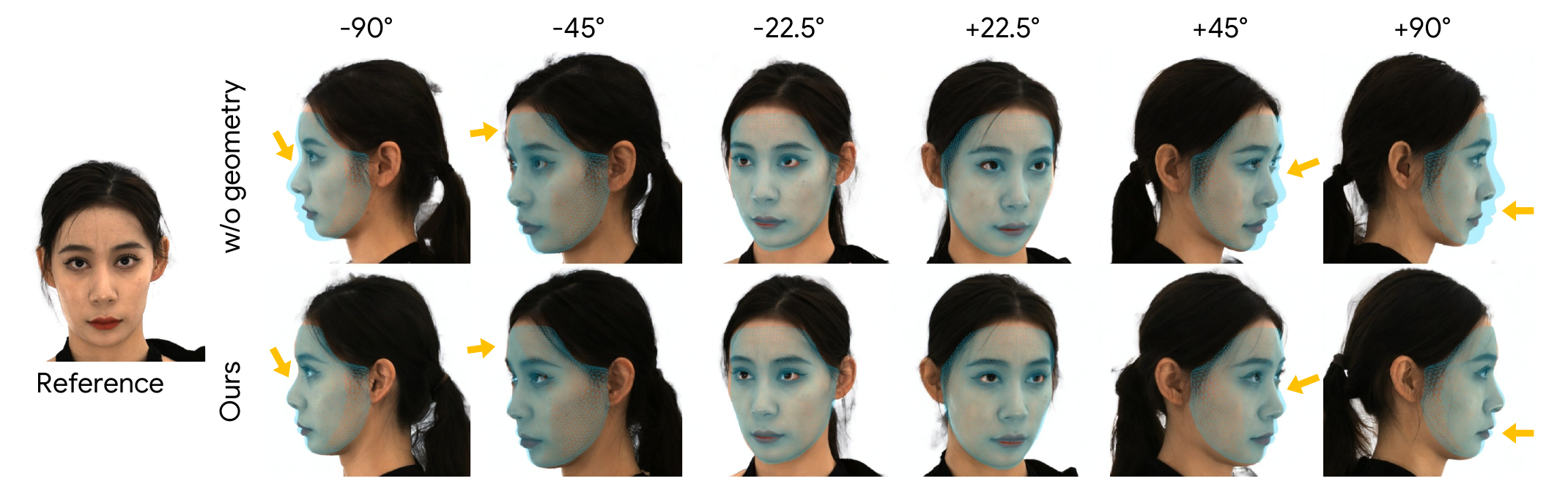}
\caption{\textbf{Qualitative ablation on the geometry stream.} Compared to the variant without geometry stream (top), \ourmodel (bottom) achieves tighter FLAME mesh alignment across all viewpoints, with particularly improved consistency at facial boundaries under large pose variations (yellow arrows).}
\label{fig:ablation}
\end{figure}

\paragraph{Effect of geometry stream.}
To evaluate the contribution of the geometry stream, we compare \ourmodel against a variant trained without the geometry stream and UV position map supervision, which corresponds to the CAT3D~\cite{Gao2024CAT3DCA} baseline trained under the same setting for 50k iterations. As shown in Table~\ref{tab:ablation}, removing the geometry stream leads to a consistent drop across all metrics, confirming that jointly generating a canonical UV position map as a shared geometric representation effectively regularizes cross-view appearance consistency, and that the improvement is not simply attributable to training iterations or data. Fig.~\ref{fig:ablation} further illustrates that the geometry stream yields tighter FLAME mesh alignment across all viewpoints, with particularly improved consistency at self-occluded regions such as the nose, forehead, and jawline under large pose variations.

\begin{table*}[t]
\centering
\caption{\textbf{Ablation study on geometry stream, token design, and alignment loss on RenderMe-360~\cite{Pan2023RenderMe360AL}.} We report averaged metrics across all viewpoints to evaluate the contribution of each design component. V2V error denotes the vertex-to-vertex distance (mm) between the generated and ground-truth FLAME mesh.}
\label{tab:ablation}
\resizebox{\linewidth}{!}{
\begin{tabular}{lcccccccc}
\toprule
\multirow{2}{*}{Model} &
\multirow{2}{*}{\begin{tabular}[c]{@{}c@{}}Geometry\\stream\end{tabular}} &
\multirow{2}{*}{\begin{tabular}[c]{@{}c@{}}Learnable\\token\end{tabular}} &
\multirow{2}{*}{\begin{tabular}[c]{@{}c@{}}Alignment loss\end{tabular}} &
\multicolumn{4}{c}{Multi-view images (6 views)} &
UV position map \\
\cmidrule(lr){5-8}
\cmidrule(lr){9-9}
 & & & & PSNR$\uparrow$ & SSIM$\uparrow$ & LPIPS$\downarrow$ & CSIM$\uparrow$ & V2V error$\downarrow$ \\ \midrule
(a) CAT3D~\cite{Gao2024CAT3DCA} & \texttimes & \texttimes & \texttimes          
& 15.12  & 0.7499  & 0.2495  & 0.7020 & - \\
(b) & \checkmark & \texttimes & \texttimes          
& 15.33  & 0.7507  & 0.2474  & 0.7069    & 15.26  \\
(c) & \checkmark & \checkmark & \texttimes          
& \underline{15.61}  & \textbf{0.7597}  & \underline{0.2359}  & \underline{0.7161} & \underline{13.12}  \\
(d) & \checkmark & \texttimes & \checkmark          
& 15.50  & 0.7520  & 0.2424  & 0.7137 &  15.00 \\ \midrule
(e) \textbf{\ourmodel (Ours)} & \checkmark & \checkmark & \checkmark   
& \textbf{15.71}  & \underline{0.7595}  & \textbf{0.2300}  & \textbf{0.7226} & \textbf{12.65}  \\
\bottomrule
\end{tabular}
}
\end{table*}

\begin{figure}[t]
  \centering
  \includegraphics[width=1\textwidth]{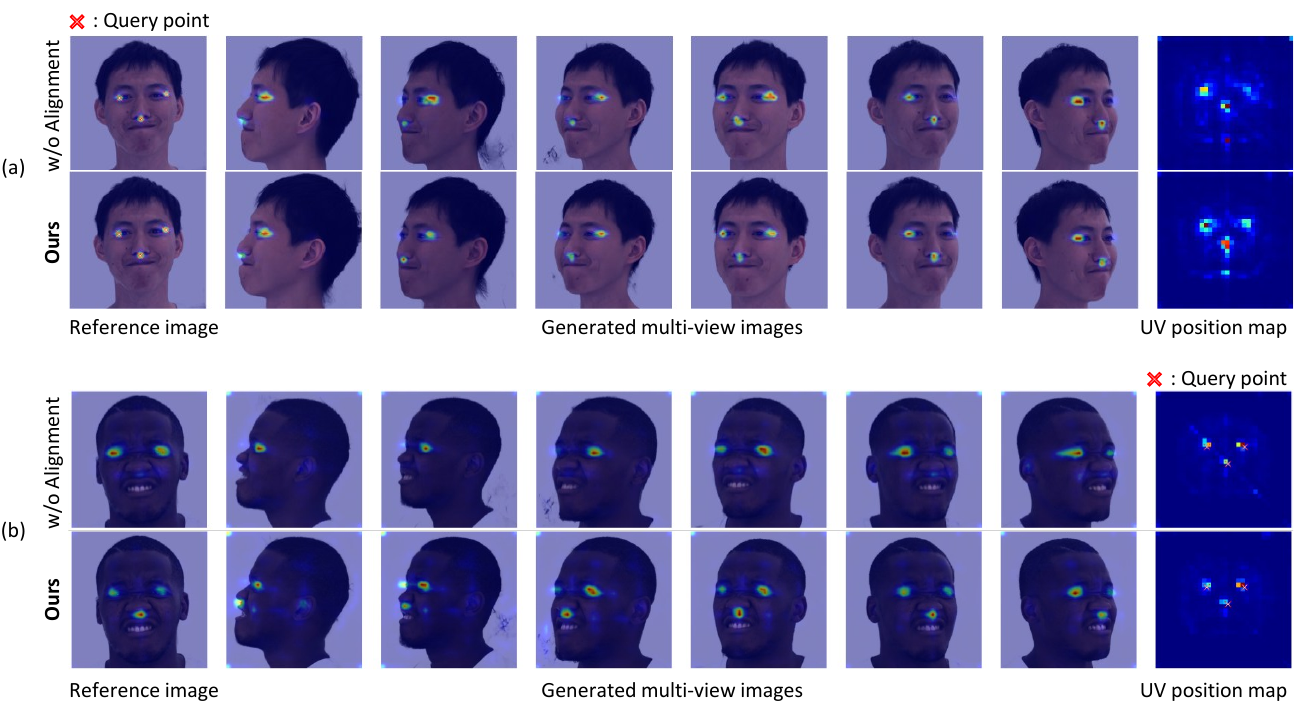}
\caption{\textbf{Qualitative ablation on the geometry-guided attention alignment loss.} (a) Cross-attention from appearance to geometry, with a query point on the reference image. (b) Cross-attention from geometry to appearance, with a query point on the UV position map. Compared to the variant without the alignment loss (top rows), GeoFace with alignment supervision (bottom rows) produces more focused and geometrically consistent cross-attention maps, leading to improved appearance generation across all viewpoints.}
\label{fig:attention_ablation}
\end{figure}

\begin{figure}[h]
  \centering
  \includegraphics[width=\linewidth]{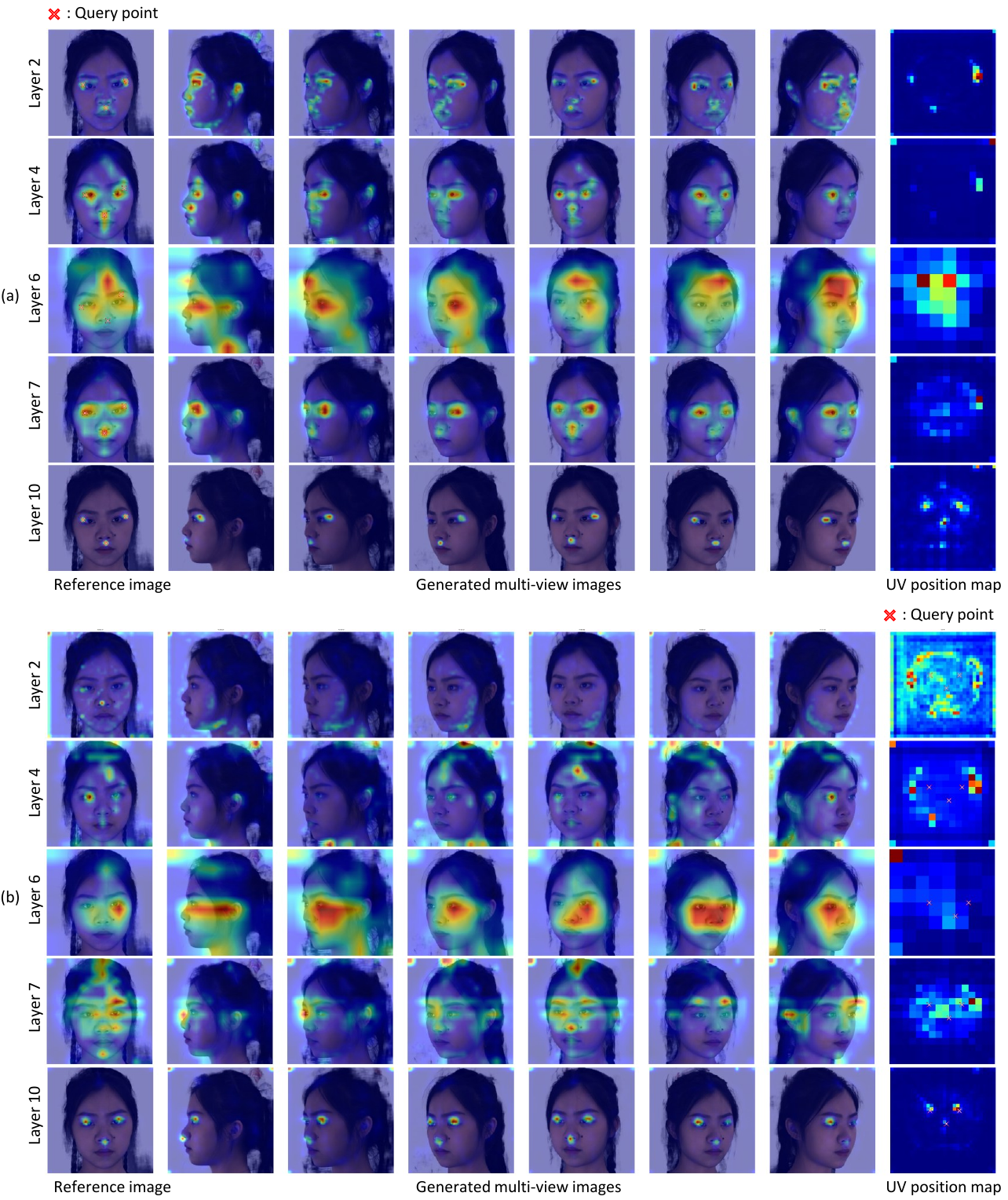}
\caption{\textbf{Layer-wise cross-attention maps between geometry and appearance streams.} (a) Cross-attention from appearance to geometry, with a query point on the reference image. (b) Cross-attention from geometry to appearance, with a query point on the UV position map. Consistent with the observation in CAMEO~\cite{kwon2025cameo}, layer 10 yields the most spatially localized correspondence in both directions.}
\label{fig:attention_layerwise}
\end{figure}

\paragraph{Effect of learnable camera token.}
To validate the design of the geometry stream conditioning, we compare our learnable camera token against a fixed ray token with the identity pose, which treats the canonical UV position map as if observed from the reference viewpoint.
However, since the canonical UV position map is defined in FLAME space and is inherently view-independent, assigning a fixed reference-view pose imposes a spurious viewpoint assumption that conflicts with the view-invariant nature of the geometry representation.
In contrast, our learnable token is optimized end-to-end, allowing the model to discover an implicit positional embedding suited for FLAME canonical space without any viewpoint bias.
As shown in Table~\ref{tab:ablation}, the identity pose leads to reduced performance across all metrics, confirming that a trainable view-agnostic conditioning is essential for stable and accurate geometry generation.

\paragraph{Effect of geometry-guided attention alignment loss.}
To validate the proposed geometry-guided attention alignment loss, we compare our full model against a variant trained without $\mathcal{L}_{\text{align}}$.
As shown in Figure~\ref{fig:attention_ablation}, without alignment supervision the cross-attention maps on the UV position map are diffuse and poorly localized, whereas our full model produces sharper attention that correctly concentrates on the corresponding facial region across all viewpoints.
Table~\ref{tab:ablation} further demonstrates that our full model achieves consistent quantitative improvements across all metrics, confirming that explicit correspondence supervision on the geometry-appearance cross-attention pathway is essential for accurate multi-view face generation.
We additionally note that, consistent with CAMEO~\cite{kwon2025cameo}, layer 10 of the UNet decoder yields the most spatially localized correspondence in both the appearance-to-geometry and geometry-to-appearance directions, as visualized in Figure~\ref{fig:attention_layerwise}, and we therefore apply supervision at this layer.

\section{Conclusion}
\label{sec:conclusion}
We present \textbf{\ourmodel}, a geometry-constrained multi-view diffusion framework that jointly generates multi-view RGB images and 3D face geometry from a single reference image. By introducing a dedicated geometry stream that denoises a canonical UV position map alongside the appearance streams, \ourmodel enforces geometric consistency as an intrinsic property of the generative process rather than a post-hoc constraint. Evaluations on RenderMe-360 and NeRSemble confirm that \ourmodel consistently outperforms existing baselines across all metrics and viewpoint ranges, with particularly pronounced gains under large pose variations. We hope this work will benefit downstream applications including 3D reconstruction, avatar creation, and immersive content generation.

\paragraph{Limitations and future directions.}
While \ourmodel successfully preserves facial identity and geometric consistency across diverse viewpoints, it builds on FLAME as the geometric backbone and may struggle to represent regions outside the facial surface, such as hair, ears, and teeth, which could limit geometry completeness for subjects with complex hairstyles or accessories. Extending the geometric representation beyond the face region and further validating generalization to in-the-wild images remain interesting directions for future work.

\clearpage

\bibliographystyle{plainnat}
\bibliography{neurips_2026}

\newpage

\end{document}